%% file: main.tex
\documentclass[letterpaper, 10 pt, conference]{ieeeconf}  %

\IEEEoverridecommandlockouts                              %

\let\labelindent\relax
\usepackage{amsmath}
\usepackage{amssymb}
\usepackage{esvect}
\usepackage{tabularx}
\usepackage{booktabs}
\usepackage{enumerate}
\usepackage{array,multirow}
\usepackage{tablefootnote}
\usepackage{enumitem}
\usepackage{caption}
\usepackage{subcaption}
\usepackage[export]{adjustbox}
\usepackage[table]{xcolor}
\usepackage{pifont}
\usepackage[T1]{fontenc}
\usepackage{soul} %

\usepackage{flushend}

\makeatletter
\let\NAT@parse\undefined
\makeatother
\usepackage[pagebackref=true,breaklinks=true,letterpaper=true,hidelinks,bookmarks=false]{hyperref}

\newcommand{\eloih}[1]{\textcolor{black}{#1}}

\newcommand{\yihongh}[1]{\textcolor{black}{#1}}

\definecolor{mylocerror}{RGB}{213, 94, 0}
\definecolor{myfn}{RGB}{0, 114, 178}
\definecolor{myfp}{RGB}{0, 158, 115}
\definecolor{myids}{RGB}{204, 121, 167}
\definecolor{myGTann}{RGB}{0, 160, 0}
\definecolor{myforecast}{RGB}{255, 165, 0}
\definecolor{myegocar}{RGB}{200, 0, 0}
\definecolor{myroadline}{RGB}{0, 190, 255}
\definecolor{mylaneboundary}{RGB}{173, 0, 255}
\definecolor{mydrivablearea}{RGB}{0, 0, 255}
\definecolor{staticcars}{RGB}{173, 173, 173}

\title{\LARGE \bf
Towards Motion Forecasting with Real-World Perception Inputs: Are End-to-End Approaches Competitive?
}

\author{Yihong Xu$^1$ Lo\"ick Chambon$^{1,2}$ \'Eloi Zablocki$^1$ Micka\"el Chen$^1$\\ Alexandre Alahi$^3$ Matthieu Cord$^{1,2}$ Patrick P\'erez$^1$
\thanks{$^{1}$ Valeo.ai, Paris, France; Email: \texttt{firstname.lastname@valeo.com}}
\thanks{$^{2}$ Sorbonne Universit\'e, Paris, France.}%
\thanks{$^{3}$ EPFL, Lausanne, Switzerland; Email: \texttt{alexandre.alahi@epfl.ch}}
\thanks{Corresponding author: Y. Xu, \texttt{yihong.xu@valeo.com}}
}

\begin{document}

\maketitle
\thispagestyle{empty}
\pagestyle{empty}

\begin{abstract}
Motion forecasting is crucial in enabling autonomous vehicles to anticipate the future trajectories of surrounding agents. To do so, it requires solving mapping, detection, tracking, and then forecasting problems, in a multi-step pipeline. In this complex system, advances in conventional forecasting methods have been made using curated data, i.e., with the assumption of perfect maps, detection, and tracking. This paradigm, however, ignores any errors from upstream modules. Meanwhile, an emerging end-to-end paradigm, that tightly integrates the perception and forecasting architectures into joint training, promises to solve this issue. However, the evaluation protocols between the two methods were so far incompatible and their comparison was not possible. In fact, conventional forecasting methods are usually not trained nor tested in real-world pipelines (e.g., with upstream detection, tracking, and mapping modules). In this work, we aim to bring forecasting models closer to the real-world deployment. First, we propose a unified evaluation pipeline for forecasting methods with real-world perception inputs, allowing us to compare conventional and end-to-end methods for the first time. Second, our in-depth study uncovers a substantial performance gap when transitioning from curated to perception-based data. In particular, we show that this gap (1) stems not only from differences in precision but also from the nature of imperfect inputs provided by perception modules, and that (2) is not trivially reduced by simply finetuning on perception outputs. Based on extensive experiments, we provide recommendations for critical areas that require improvement and guidance towards more robust motion forecasting in the real world. The evaluation library for benchmarking models under standardized and practical conditions is provided: \url{https://github.com/valeoai/MFEval}.
\end{abstract}

\section{Introduction}
\label{sec:intro}

\begin{figure}[t]
\centering
\begin{subfigure}{0.325\linewidth}
\includegraphics[trim={0 0 0 2.5cm},clip,width=\linewidth]{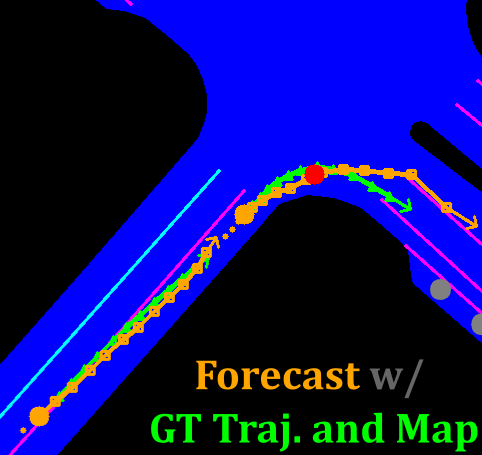} 
  \vspace{-0.6cm}
  \caption{\vspace{-0.2cm}}
  \label{fig:gt_in_gt_map}
\end{subfigure}
\hfill
\begin{subfigure}{0.325\linewidth}
\includegraphics[trim={0 0 0 2.5cm},clip,width=\linewidth]{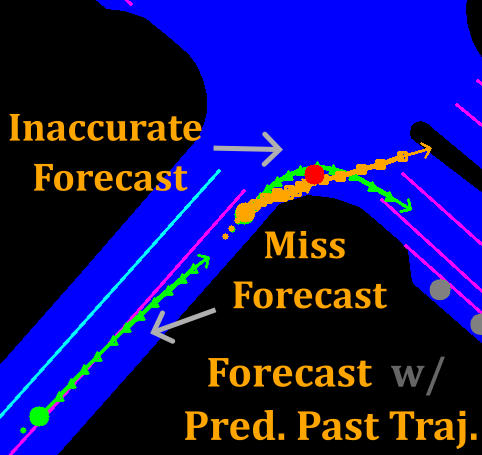}
   \vspace{-0.6cm} 
  \caption{\vspace{-0.2cm}}
  \label{fig:noisy_in_gt_map}
\end{subfigure}
\hfill
\begin{subfigure}{0.325\linewidth}
\includegraphics[trim={0 0 0 2.5cm},clip,width=\linewidth]{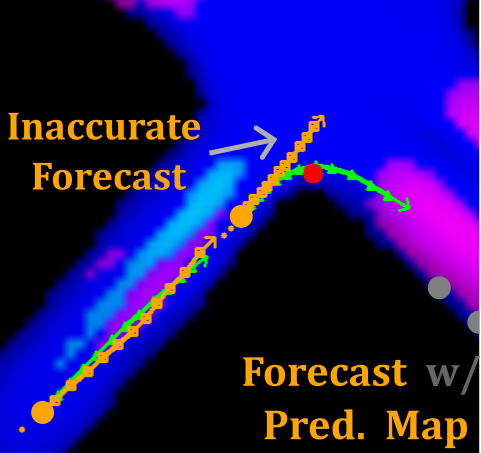}
   \vspace{-0.6cm} 
  \caption{\vspace{-0.2cm}}
  \label{fig:gt_in_noisy_map}
\end{subfigure}
\caption{
{\textbf{Issues of deploying forecasting models to the real world}}.
We show in a nuScenes example \cite{nuscenes} the {forecasts} (in orange) inferred by a motion forecasting model \cite{yuan2021agentformer} compared to {ground-truth annotations} (in green), the {ego car location} (in red) and the {static vehicles} (in gray) on predicted or curated maps. 
{(a) Satisfying forecasting performance in a curated setting; (b) When past trajectories are inferred from tracking models \cite{zhu2019class, Weng2020_AB3DMOT}, an agent is not detected and the forecasting model yields poor predictions; (c) When the map is inferred online \cite{harley2022simple}, the forecasting model does not anticipate the future turn of one agent. %
}
}
\end{figure}

Motion forecasting plays an important role for autonomous vehicles (i.e., ego vehicles), enabling them to anticipate future trajectories of agents in their surroundings (i.e., vehicles of interest) and, accordingly, to plan safely \cite{nishimura2022rap,cao2022robust_trajectory_adversarial,li2021risk_aware_planner}.
{This complex task is usually shared between upstream modules for mapping, detecting, and tracking agents, and the forecasting module proper.
In this system,} most \textit{conventional} forecasting works
\cite{covernet,salzmann2020trajectron++,yuan2021agentformer,densetnt, multipath,multipath++,mfp,mtp,mosa,cui2023gorela,wang2023ganet,nayakanti2023wayformer} set themselves in a setting with perfectly solved upstream tasks, and are trained primarily using inputs from curated offline annotations, including clean agent past trajectories and detailed road information \cite{nuscenes}.
A forecast with such curated inputs is shown in \autoref{fig:gt_in_gt_map}.
However, when deployed in real-world settings, motion forecasting modules rely on data provided by upstream detection, tracking and mapping modules \cite{yuan2022keypoints, zhang2022mutr3d, chen2023voxelnext, yin2021center, Weng2020_AB3DMOT}, often resulting in lower-quality predictions compared to the curated datasets used in research publications \cite{okumura2016challenges_perception}.
{As an example, when the inputs (the past trajectories or the map) are degraded (in \autoref{fig:noisy_in_gt_map} or \autoref{fig:gt_in_noisy_map} respectively), the predictions become worse compared to \autoref{fig:gt_in_gt_map}, e.g., failing to anticipate well the turn (b and c in the figure) and to simply forecast an agent that is not detected (b in the figure).}

As an alternative to this conventional pipeline of a perception model followed by a forecasting model, \textit{end-to-end} methods \cite{gu2023vip3d, uniad} have recently received some attention. They advocate for joint perception-forecast training and inference with a more tightly integrated architecture, typically only using perception outputs as an intermediary representation or a multi-task training objective.
Yet, both paradigms have not been compared.
In fact, conventional forecasting models are not usually designed nor evaluated jointly with upstream perception models, and it is not known how they perform when integrated into the deployed pipeline.

In this work, our objective is to bring forecasting models closer to real-world deployment.
Accordingly, we first design a unified evaluation protocol for forecasting by integrating the upstream perception modules into the conventional forecasting evaluation.
Second, this benchmark enables us to assess and compare end-to-end methods with conventional pipelines, which was previously not feasible.
Third, we also uncover a substantial drop in performance when transitioning from the curated setting to a real-world scenario. %
\begin{figure*}[t]
\centering\includegraphics[width=\linewidth]{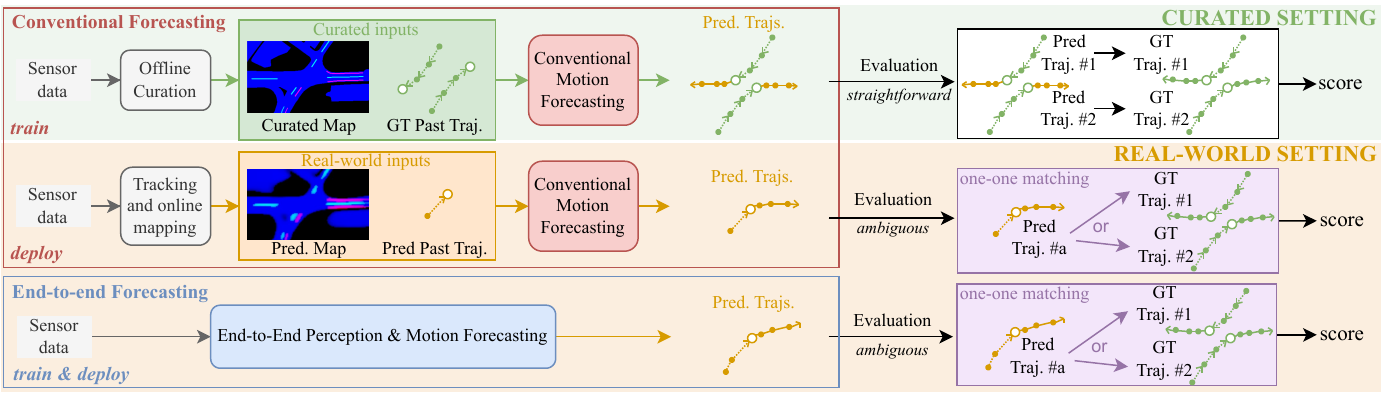}
  \caption{\textbf{Study overview}. We study the challenges of deploying motion forecasting models into the real world when only predicted perception inputs are available. We compare (\autoref{sec.eval}): (1) {(\textit{top}) `conventional methods' \cite{yuan2021agentformer,kim2021lapred} (i.e., methods trained on curated input data) where (middle) we} directly replace the curated inputs with real-world data, and (2) (\textit{bottom}) `end-to-end methods' \cite{gu2023vip3d,uniad} that are trained and used with perception modules.
  {In the real-world setting, evaluation is challenging as the past tracks are estimated with arbitrary identities, making it difficult to establish a direct correspondence to GT identities. Therefore, we propose a matching process (purple) to assign predictions to GT and thus evaluate forecasting performances (\autoref{sec.eval}).} %
  Moreover, we study in depth the impact changing from {curated data} (green) to {real-world} (orange) mapping (\autoref{sec:map}), or detection and tracking (\autoref{sec:det_track}) errors to motion forecasting.}
\label{fig:pipeline}
\end{figure*}
While it may seem intuitive that a gap exists, the issue has mostly been overlooked in driving contexts, hence, it has never been properly measured and thoroughly characterized.
We then conduct a complete study to identify a wide range of obstacles that cause this gap, and we demonstrate as not easily solved.
In particular, our study covers the impacts of using state-of-the-art perception modules for detection and tracking \cite{zhu2019class, Weng2020_AB3DMOT,zhang2022mutr3d,yin2021center,chen2023voxelnext,gu2023vip3d,uniad}, and online rasterized or vectorized mapping \cite{harley2022simple, bartoccioni2022lara, MapTR}, on the performance of different motion forecasting models \cite{yuan2021agentformer, kim2021lapred}.

{As summarized in \autoref{fig:pipeline}: (1) This work provides an evaluation protocol to benchmark forecasting models with real-world inputs (\autoref{sec.eval}); (2) This benchmark allows us to compare, for the first time, the end-to-end and conventional models under standardized and practical conditions (\autoref{sec.eval} and \autoref{sec:map}); (3) Being a missing brick in current literature, our extensive experiments on the impact of various perception errors (detection, tracking, and mapping) on forecasting shed light on the critical areas that need improvement (\autoref{sec:map} and \autoref{sec:det_track}); (4) Based on the findings, we provide recommendations towards robust motion forecasting in the real world (\autoref{sec:discussion}).} 

\section{Related Work}
\label{sec:rw}
\textbf{Conventional motion forecasting} methods
\cite{covernet,salzmann2020trajectron++,yuan2021agentformer,densetnt, multipath,multipath++,mfp,mtp,mosa,cui2023gorela,wang2023ganet,nayakanti2023wayformer} focus on better leveraging {and modeling the interactions between their diverse inputs.} %
For example, AgentFormer \cite{yuan2021agentformer} formulates the forecasting problem as the modeling of all surrounding agents conditioned on their past trajectories and contextual road elements. LaPred \cite{kim2021lapred} predicts per-agent future trajectories leveraging the closest {vectorized} lane information and past trajectories of neighboring agents. The lane and agent information is combined to enforce the use of {vectorized} map information and the behavior of neighbor agents, building a strong baseline even with a simple yet efficient MLP-based trajectory decoder. We assess the performance of the conventional methods by substituting the curated data by real-world inputs and {highlight the open} challenges.

\textbf{Object tracking and online mapping.}
Motion forecasting takes as inputs agents' past trajectories and map information. In the real-world setting, they are inferred respectively by motion trackers and online mappers. Motion tracking jointly performs object detection and association with different modalities. For example, MUTR3D \cite{zhang2022mutr3d} utilizes a transformer architecture with 3D track queries to model spatial and appearance coherence across multiple cameras and frames. CenterPoint \cite{yin2021center} leverages LiDAR information to first detect the centers of objects and then regress other attributes (size, orientation, etc.). Similarly, VoxelNext \cite{chen2023voxelnext} directly detects and tracks objects in point-cloud based on sparse voxel features. For mapping, curated maps are in general costly and hard to maintain. As an alternative, some works have proposed to estimate them online, first from a single front camera \cite{can2021structured,can2022understanding}.
Then, the development of bird's-eye view (BEV) representations led to the use of surround RGB cameras \cite{liftsplat, MapTR,bartoccioni2022lara,harley2022simple} optionally coupled with LiDAR point-clouds acquisitions \cite{li2021hdmapnet, vectormapnet, MapTR, sun2023exploring}. Our work studies the impact of predicted trajectories and online mapping compared to the curated settings most motion forecasting models rely on. %

\textbf{Motion forecasting in end-to-end driving pipelines.}
    Recent works propose to learn forecasting models directly from raw data \cite{fast_and_furious,liang2020pnpnet,liftsplat, fiery,weng2022affinipred,zhang2022trajectory,mtp,gu2023vip3d,uniad}. To address the issue of error propagation in the downstream forecasting module, MTP \cite{mtp} and FutureDet \cite{peri2022forecasting_future_detection} replace the one-to-one assignment in tracking with a one-to-many one, as motion forecasting performances can be deteriorated by identity switches and detection errors \cite{mtp}.
    AffiniPred \cite{weng2022affinipred} and Zhang et al. \cite{zhang2022trajectory} perform implicit data association by using detections and their affinity matrices as inputs instead of working on past trajectories. These studies have a special focus on the tracking error while {\cite{bahari2022vehicle} and \cite{sarva2023advd} tackle a subset of imperfections with an adversarial scene or object generation. Differently, we focus on understanding the impacts of real-world inputs from various state-of-the-art perception methods on the different motion forecasting paradigms.} Yet, joint perception-to-forecasting models (ViP3D \cite{gu2023vip3d}, UniAD \cite{uniad}) have not been directly compared to the established pure motion forecasting models, primarily due to differences in their approaches and evaluation criteria.
    An adapted evaluation protocol is needed with metrics considering the upstream errors \cite{philion2020planner_centric_metrics, weng2023joint, peri2022forecasting_future_detection, ivanovic2022heterogeneous}.
    Our work is the first to pull both approaches into a single evaluation framework.
    
\section{A Unified Evaluation Benchmark}
\label{sec.eval}

In our study, we analyze conventional forecasting models {when they} are confronted, instead of the curated data, to {outputs of} various state-of-the-art perception modules for detection and tracking \cite{zhu2019class, Weng2020_AB3DMOT,zhang2022mutr3d,yin2021center,chen2023voxelnext,gu2023vip3d,uniad}, and to online rasterized or vectorized mapping \cite{harley2022simple, bartoccioni2022lara, MapTR} on nuScenes dataset \cite{nuscenes}.
As illustrated in \autoref{fig:pipeline}, we replace the curated annotations with real-world predictions as inputs to conventional motion forecasting models \cite{yuan2021agentformer, kim2021lapred} and examine the forecasting and perception performance. 
By doing so, we can incorporate the same predicted perception inputs as in end-to-end approaches \cite{gu2023vip3d, uniad}.
However, the comparison is confronted with {challenges that we first detail and then address {here.}
As a direct application, we are able to compare, for the first time, recent methods in the end-to-end forecasting paradigm to conventional methods.

\subsection{The need for matching}
Standard forecasting datasets, such as the used nuScenes dataset \cite{nuscenes}, provide past information about the ground-truth agents, including their number, identities, and trajectories.
Conventional forecasting models rely on these identities to train their models and compute their scores.
However, in the real world, detected agents are not inherently linked to ground-truth agents because GT identities are not provided during inference, and arbitrary identities are assigned during tracking.
To address this, a one-to-one matching is needed to assign predictions to ground-truth agents.
Similar to the multi-object tracking problem \cite{bernardin2008evaluating}, we use Hungarian matching with a matching threshold of 2 meters between object center L2 distances at the starting frame for forecasting, i.e., $t = 0$.

\subsection{The need for suitable metrics}
{For similar reasons, standard forecasting metrics --- $\textbf{minADE}_k$, the minimum over $k$ predictions of Average Distance Error (the average of point-wise L2 distances between the prediction and ground-truth forecasts), $\textbf{minFDE}_k$, the minimum over $k$ predictions of Final Distance Error (the L2 distance at the final future time step),
and $\textbf{MR}_{k@x}$, MissRate, the ratio of forecasts having $\textbf{minFDE}_k$ > $x$ = 4 meters --- are built on the assumption that the models have matching identities from past and future tracks. These metrics solely consider the forecasting quality of \emph{matched} prediction-GT pairs, without penalizing missed or falsely predicted agents.

{As this does not provide the full picture of real-world forecasting performance, we} propose to consider the Mean Forecasting Average Precision ($\text{mAP}_{f}$) \cite{peri2022forecasting_future_detection} that shares the same formulation as detection AP \cite{lin2014microsoft}. However, $\text{AP}_{f}$ considers as false positives not only trajectories with incorrect first-frame detections (center L2 distance at $t = 0$ bigger than 2m), but also those having correct first-frame detections but inaccurate forecasts ($\textbf{minFDE}_k$ > 4m) .
The APs are then averaged over the classes `car', `truck', and `bus' -- $\text{mAP}_{f}$. Unlike \cite{gu2023vip3d, uniad}, to reflect the forecasting quality, we only evaluate ground-truth (GT) \textit{vehicle} agents having full \emph{moving} future trajectories as in nuScenes Prediction challenge, resulting in different and more realistic performance than the one shown in \cite{gu2023vip3d, uniad} that consider mostly static objects including pedestrians.
\subsection{Conventional vs.\ End-to-End forecasting}
\input{tabs/vip3d_vs_ag_vs_uniad}
{Recent end-to-end methods \cite{gu2023vip3d, uniad}, though providing a promising direction, are poorly evaluated (e.g., including static objects), which hinders the understanding of their underlying issues.
Since both paradigms have never been fairly compared, we make the first step to compare them with the same perception inputs from end-to-end pipelines.}

We consider AgentFormer \cite{yuan2021agentformer}, using rasterized maps, and LaPred \cite{kim2021lapred}, using vectorized maps, as strong representatives of conventional methods. %
For the end-to-end paradigm, we choose two state-of-the-art methods, namely, ViP3D \cite{gu2023vip3d}, an end-to-end motion forecasting model, and UniAD \cite{uniad}, an end-to-end forecasting and planning model. The choice is made considering the availability of code and their distinct structures. {Please refer to the related works for more details.}

As shown in \autoref{tab:ag_vs_all}, {{the first} observation is that \emph{recent state-of-the-art end-to-end models do not exhibit superior performance in forecasting, compared to conventional methods that have been trained only with curated inputs.}} Precisely, with the {detection and tracking} inputs from ViP3D, AgentFormer outperforms ViP3D significantly (more than 2 times higher $\text{mAP}_{f}$) without being jointly trained with such perception inputs.
Since AgentFormer combines agent past history and inter-agent interactions in an attention module, without explicitly separating different trajectories, %
this joint interaction may help to resist the detection and tracking errors. Similarly, \yihongh{with no finetuning, LaPred leads the performance in $\text{mAP}_{f}$ with a simpler trajectory predictor, compared to UniAD with its end-point attention-based refinement and  physically-based kinematic model.} Lastly, we observe that end-to-end pipelines (UniAD and ViP3D), despite operating in a much more realistic setting, are still \textit{very far} from conventional forecasting methods with curated data (i.e., AgentFormer and LaPred with ground-truth inputs). {The observations imply that it might not be trivial to train the end-to-end model jointly with perception modules without tackling their errors in downstream forecasting.} 

\input{tabs/with_wo_map}
\section{Impacts of the Map Quality}
\label{sec:map}
We now study in more detail how conventional forecasting methods use their inputs and if we can replace them with real-world perceptions. In this section, we intervene on the map input of AgentFormer \cite{yuan2021agentformer}, LaPred \cite{kim2021lapred}, ViP3D \cite{gu2023vip3d} and UniAD \cite{uniad}. Originally, AgentFormer uses \emph{rasterized} curated maps; \yihongh{LaPred and ViP3D both leverage \emph{vectorized} curated maps}, while UniAD uses online mapping inferred from camera data.
\subsection{Removing map information entirely}

{Our first goal is to assess the dependence of forecasting performances to map information.}
To do so, we replace the maps at the input of the forecasting modules with empty {ones}.
In practice, we consider two distinct scenarios: 1) {a direct} `\emph{transfer}' without finetuning on empty maps, and 2) a `\emph{finetune}' one where each model uses empty maps throughout both finetuning and inference stages.

From \autoref{tab:with_wo_map}, our experiments reveal that \emph{the presence or absence of maps has minimal impact on the performance of the end-to-end ViP3D and UniAD models.}
For ViP3D, training the model with curated maps and replacing the map with an empty map during inference shows negligible change in results (\emph{transfer} setting).
Moreover, finetuning ViP3D without map information leads to slightly improved performance on the validation set (\emph{finetune} setting), indicating that the map is not utilized at all in ViP3D.
For UniAD, the model exhibits a slight performance drop when using an empty map during inference, but finetuning without map information yields similar results to using the online map, suggesting that it is not well utilized in UniAD either. On the other hand, for AgentFormer and LaPred, finetuning without maps does not close the performance gap compared to having maps, indicating their better utilization of contextual information.
Although the poor usage of maps {has already been pointed out by} previous work \cite{benyounes2022cab} on some conventional methods \cite{salzmann2020trajectron++, halentnet}, \textit{we show for the first time that the issue persists in both recent end-to-end models \cite{uniad} and \cite{gu2023vip3d}.} In the following, we study the impact of different aspects of the map in more detail. As we have determined that ViP3D and UniAD do not effectively utilize the map, our following analysis concentrates on AgentFormer and LaPred.

\subsection{From curated to online mapping}
We now assess the performance gap when going from curated maps to using the output of an online mapping method.
We consider three different mapping methods: SimpleBeV \cite{harley2022simple} and LaRa \cite{bartoccioni2022lara} for rasterized BEV map prediction and MapTR \cite{MapTR}, an online mapping method predicting vectorized map elements. LaRa obtains 0.361 mIoU and 0.458 for SimpleBeV. MapTR exhibits state-of-the-art performance with 62.8 mapping mAP \cite{MapTR}.

In \autoref{tab:with_wo_map}, we observe that for direct \emph{transfer}, SimpleBeV and LaRa maps perform worse than empty maps, indicating a significant domain gap as the forecasting model is unable to use online predictions directly. Finetuning improves performance for both empty and online maps, but the latter, although better, still underperforms curated maps. Thus, online maps are so far insufficient to replace curated maps. One issue could be their limited range around the ego car, much smaller than curated maps. Additionally, we find that high-level map elements such as `lane', `drivable area' (0.388\,$\rightarrow$\,0.296 mAP$_f$ for AgentFormer, and 0.760\,$\rightarrow$\,0.611 for LaPred) are more impactful than detailed map information (`road' and `lane dividers') that could be better leveraged (0.388\,$\rightarrow$\,0.337 mAP$_f$ for AgentFormer and 0.760\,$\rightarrow$\,0.668 for LaPred).

\section{Impacts of Detection and Tracking}
\label{sec:det_track}
\input{tabs/my_real_world_inputs}
{In this section, we study the past trajectories inputs of conventional forecasting models} \cite{yuan2021agentformer,kim2021lapred},
by replacing GT past trajectories with outputs of real-world tracking models (\autoref{sec:tracking:real-world}) or by artificially intervening on them (\autoref{subsec:syntheticbreakdown}). {This allows us to study the importance of precise agents' positioning and identification.} We also study in \autoref{sec:finetune} how much can simple finetuning can improve the forecasting models.
To quantify the perception input quality, we count the number of false positives (FP), i.e., predicted objects not associated with any GT object, of false negatives (FN), i.e., GT objects not associated with any prediction, and of identity switches (IDS), i.e., assigning wrongly the detection to an agent of a different identity. %
For ease of interpretation, FP, FN and IDS are combined into Multiple Object Tracking Accuracy (MOTA) \cite{bernardin2008evaluating}. We also compute Multiple Object Tracking Precision (MOTP), quantifying the average positional accuracy over matched objects. Details can be found in \cite{bernardin2008evaluating}.

\subsection{From curated to predicted agents}
\label{sec:tracking:real-world}
We investigate how forecasting models would react in a realistic setting where the agents' positions and identities are not curated but predicted by perception models.
Our selection of such models includes recent state-of-the-art methods that can be LiDAR-based, such as MegVii \cite{zhu2019class} + AB3DMOT \cite{Weng2020_AB3DMOT}, CenterPoint \cite{yin2021center} and VoxelNext \cite{chen2023voxelnext}, or camera-based such as MUTR3D \cite{zhang2022mutr3d} with either ResNet-50 (R50) or ResNet-101 (R101) backbones, ViP3D \cite{gu2023vip3d}, and UniAD \cite{uniad}.
In \autoref{tab:real_world_results}, we also show results obtained with the curated GT as inputs for comparison.
First, we observe that using real-world inputs leads to a very significant forecasting performance drop for both AgentFormer (0.343\,$\rightarrow$\,0.112 in mAP$_f$) and LaPred (0.757\,$\rightarrow$\,0.317). 
Second, while LiDAR-based methods are much better at detection and tracking than camera-based methods, we remark that this only translates into marginally better forecasting scores. %
For instance, ViP3D Det\&Track even manages to achieve better MR and comparable minFDE to LiDAR-based methods, while UniAD Det\&Track is not so far behind.
Then, we can observe that replacing ground-truth identities of agents with outputs of a tracking model from 
 \cite{yin2021center} does not significantly degrade the performance (e.g., 0.343\,$\rightarrow$\,0.317 in mAP$_f$ for AgentFormer) despite its worse IDS. This indicates that IDS in near agents is not as impactful as detection errors.
\subsection{Detection and tracking errors breakdown}
\label{subsec:syntheticbreakdown}
\input{tabs/det_track_noise}
A significant performance drop is observed using real-world inputs that contain a complex combination of highly correlated errors. To give insights on which type of errors dominates, we break down the errors that occur in the \textit{past trajectories} into four types: FP by duplicating and perturbing ground-truth agents' locations within 5 meters; FN (except at t$=$0, i.e., the starting point of forecasting) by randomly removing ground-truth agents; localization error (Loc. Error correctly detected but misplaced) by perturbations of ground truth within 2 meters and, IDS with nearby neighbors within 5 meters. The choice of distance is based on real-world conditions (e.g., a vehicle tends to switch identities with nearby vehicles).
We simulate errors with synthetic perturbations on the curated annotations, in varying proportions, and plot the performance of AgentFormer and LaPred in these settings in \autoref{fig:det_track_noise}. We note that for FN, we do not consider missing detections at t$=$0 since it will obviously cause catastrophic miss forecasts. Besides, we observe that localization errors are the most impactful on forecasting metrics while IDS errors have less impact, in complement with \cite{mtp, weng2022affinipred}.

\subsection{Finetuning with imperfect inputs} \label{sec:finetune}
Intuitively, one can think that the performance gap can be easily closed by finetuning. This experiment has already been conducted for map imperfections in \autoref{tab:with_wo_map} (\textit{`finetune'}) where we show that the gap persists. We further conduct finetuning experiments on the stronger model (LaPred) with (1) data augmentation by randomly adding 30\% GT past position perturbation (i.e., simulating Loc. Error). We also tried with 10\% and 50\%, yielding worse results. (2) real-world tracking results of different modalities (LiDAR-based VoxelNext \cite{chen2023voxelnext}, end-to-end UniAD \cite{uniad} and camera-based MUTR3D R101 \cite{zhang2022mutr3d}). We show in \autoref{tab:train_imperfections} that finetuning with data augmentation has very limited help in improving forecasting performance with real-world inputs. A systematic but slight improvement is observed by finetuning {with the actual} real-world inputs (on the trainset) but the improved performance is not comparable with GT annotations.

The reasons are: (1) The distribution of perturbation in the trainset is different from the one in the validation set. (e.g., MapTR obtained 94.7 mapping mAP in the trainset vs.\ 62.8 in validation); (2) The forecasting models are not designed to handle such errors; (3) Unlike weather or road domain gap \cite{lee2022learning, sun2022shift}, the map elements or the detections are often missing and simply finetuning with real-world inputs brings no benefits. This indicates that poor forecasting performance due to upstream perception errors cannot be easily fixed with basic domain adaptation methods.
\input{tabs/training_with_noise}

\subsection{Impact of the distance to ego vehicle}
\label{sec:egoagentdist}
To better characterize the differences between different detection tracking methods, we group the agents per distance to the ego vehicle and report both tracking and forecasting performances in \autoref{fig:dist_ego_car}.
First, we observe that LiDAR-based trackers fare better in general. This advantage can be explained by the fact that LiDAR sensors keep good precision regardless of range. While the superiority is especially {revealed} with MOTP, 
the relatively similar MOTA scores indicate that LiDAR-based methods struggle nearly as much as camera-based trackers for detecting far-away agents.
The superiority in MOTP however reflects that once detected, they are much better at precisely locating them.
LiDAR-based methods are better in $\text{mAP}_{f}$ despite similar MOTA is also compatible with our previous observation that the localization precision (in MOTP) is crucial for forecasting. 
This observation is significant because the ego vehicle position is often ignored in motion forecasting as the prediction {and evaluation are} usually done in an agent-centric view. %
In light of these findings, motion forecasting models and evaluation should consider agent-ego distance.

\input{tabs/ego_car}

\section{Conclusion}
\label{sec:discussion}
This work brings conventional and end-to-end methods into a joint evaluation protocol representative of real-world constraints and enables the study of diverse forecasting methods {(4 of them in the experiments)} when facing outputs of {trackers (7 considered)}, and online mapping methods {(3 considered, vectorized or rasterized, using LiDAR or camera)}.
This comparison sheds light on the poor performance of end-to-end methods and highlights the challenges that arise when interfacing perception and forecasting models. The main findings and corresponding recommendations are: 
\begin{itemize}[leftmargin=*]
    \item (\autoref{sec.eval} and \autoref{sec:map}) \emph{{The emerging `end-to-end forecasting' paradigm is so far not better than the conventional one, even in a real-world setting without finetuning. Besides, end-to-end models do not utilize map information.}} A better multi-task learning strategy and map integration design is needed to advance the end-to-end paradigm.
    \item (\autoref{sec:map} and \autoref{sec:det_track}) \emph{There is a {large and} systematic {performance} gap going from curated annotations to perception predictions, which is not reduced by simple techniques,} requiring more effort than just joint training. Also, for perception tasks, precise localization should be considered along with detection mAP or tracking accuracy and the perception range should be enlarged to a physically achievable range. %
    
    \item (\autoref{sec:det_track}) \emph{We show that the perception and forecasting quality depends on the agent-ego distance.} While intuitive once formulated, this information is missing in current benchmarks, which should be included by stratifying the evaluation according to the distance to ego-vehicle.
\end{itemize}

Finally, we encourage the community to publicly release codes and models to benchmark them with a broader spectrum of driving scenes in the future. {The advantages of end-to-end models, such as a single-loop training and easier deployment, motivate us to further investigate and improve this promising paradigm. We also note that our study can be further extended to the downstream task of motion planning.
\section*{ACKNOWLEDGMENT}
This work was supported by the ANR MultiTrans (ANR-21-CE23-0032) and CINES (HORS DARI N°A0141014181). This research received the support of EXA4MIND project, funded by a European Union´s Horizon Europe Research and Innovation Programme, under Grant Agreement N°101092944 views and opinions expressed are however those of the author(s) only and do not necessarily reflect those of the European Union or the European Commission. Neither the European Union nor the granting authority can be held responsible for them. Finally, we want to thank Florent Bartoccioni and Kaouther Messaoud for helpful discussions.
{\small
\bibliographystyle{IEEEtran}
\bibliography{biblio}
}
\end{document}

%% file: tabs/vip3d_vs_ag_vs_uniad.tex
\begin{table}[t]
\centering
    \caption{\yihongh{\textbf{Comparison of end-to-end and conventional forecasting methods} given the same detection and tracking (`Det\&Track') inputs; $k$, the number of possible forecasts, i.e., \textit{modes}.}}  %
    \resizebox{\linewidth}{!}{
    \begin{tabular}{@{}l @{\hspace{2mm}} l @{\hspace{2mm}} c @{\hspace{2mm}} c @{\hspace{2mm}} c @{\hspace{2mm}} c@{}}
    \toprule
     \textbf{Det\&Track Input} & \textbf{Forecast Method} & \textbf{mAP${_f}$} \hspace{-2mm}  $\uparrow$ & \textbf{minADE} \hspace{-2mm} $\downarrow$  & \textbf{minFDE} \hspace{-2mm} $\downarrow$  & \textbf{MR} \hspace{-2mm} $\downarrow$ \\

    \midrule
     \multirow{2}{*}{Ground truth} & AgentFormer ($k$ = 5) &0.388	&1.851	&3.875	&0.315   \\ 
     & \eloih{LaPred ($k$ = 5)} & \eloih{\textbf{0.588}}	& \eloih{\textbf{1.547}}	& \eloih{\textbf{3.176}}	& \eloih{\textbf{0.208}}   \\  
      \midrule  
                                                                           \multirow{5}{*}{\parbox[c]{2.5cm}{ViP3D (CVPR'23) \\ \cite{gu2023vip3d}}} & AgentFormer ($k$ = 5)       & {0.056}	&{2.416}	& {4.404}	&{0.353} \\
                                                                           & \eloih{LaPred ($k$ = 5)}  & \eloih{0.092}	& \eloih{2.612}	&\eloih{4.520}	&\eloih{0.282} \\
                                                                            & ViP3D ($k$ = 5)                     &0.021	&4.018	&7.040	&0.505 \\
                                                                           & \eloih{LaPred ($k$ = 6)} &  \eloih{\textbf{0.113}} & \eloih{\textbf{2.365}} & \eloih{\textbf{3.900}} & \eloih{\textbf{0.224}} \\
                                                                           & ViP3D ($k$ = 6)                     &0.034	&3.540	&5.943	&0.432\\
     \midrule  

                                                                           \multirow{5}{*}{\parbox[c]{2.5cm}{UniAD (CVPR'23) \\ \cite{uniad}}} & AgentFormer ($k$ = 5)        &0.069	 &2.530	 &4.613	  &0.384 \\
                                                                           & \eloih{LaPred ($k$ = 5)}  & \eloih{0.123} & \eloih{2.684} & \eloih{4.678}	& \eloih{0.278} \\
                                                                            & UniAD  ($k$ = 5)                     &0.094	&2.071	&3.810	&0.283  \\
                                                                           & \eloih{LaPred ($k$ = 6)} &  \eloih{\textbf{0.143}} & \eloih{2.499} & \eloih{4.212} & \eloih{0.237} \\
                                                                            & UniAD  ($k$ = 6)                   & {0.117}	&\textbf{1.842}	&\textbf{3.258}	&\textbf{0.228}  \\
     
    \bottomrule
    \end{tabular}
    }
  \label{tab:ag_vs_all}
\end{table}

%% file: tabs/with_wo_map.tex
\begin{table}[t]
\centering
    \caption{\textbf{Performance of conventional and end-to-end methods with various types of input maps.} Map is a ground-truth curated map, an empty map or an online (rasterized or vectorized) map. In the two latter cases, the model is evaluated on the new map type either directly (`\textit{transfer}') or after finetuning (`\textit{finetune}').}
    \resizebox{\linewidth}{!}{
    \begin{tabular}{@{}l l @{\hspace{2mm}} l c @{\hspace{1mm}} c @{\hspace{1mm}} c @{\hspace{1mm}} c@{\hspace{1mm}}}
    \toprule
     \textbf{Method} & \textbf{Map} & \textbf{Setting} & \textbf{mAP${_f}$} \hspace{-2mm} $\uparrow$ & \textbf{minADE} \hspace{-2mm} $\downarrow$  & \textbf{minFDE} \hspace{-2mm} $\downarrow$  & \textbf{MR} \hspace{-2mm} $\downarrow$ \\

     \midrule  
     \multirow{7}{*}{\parbox[c]{1cm}{AgentFormer~\cite{yuan2021agentformer}}} & ground truth & default                           &\textbf{0.388}	&\textbf{1.851}	&\textbf{3.875}	&\textbf{0.315}   \\  
     & empty map & \textit{transfer}                       &0.057	&2.747	&6.165	&0.668\\
     & empty map & \textit{finetune}             &0.289	&1.966	&4.221	&0.398\\
     & LaRa \cite{bartoccioni2022lara} & \textit{transfer}   &  {0.028}	& {3.246}	&{7.358}	&{0.750}\\
     (map: Raster) & LaRa \cite{bartoccioni2022lara} & \textit{finetune}    &{0.341} &{1.916}	& {4.085}	& {0.350} \\

     & SimpleBeV \cite{harley2022simple} & \textit{transfer}   & {0.034}	&{3.128}	&{7.067}	&{0.727}\\
     & SimpleBeV \cite{harley2022simple} & \textit{finetune}    &{0.361} &{1.856}	&{3.935}	&{0.333} \\

     \midrule  
     \multirow{5}{*}{LaPred \cite{kim2021lapred} } & ground truth & default                           &\textbf{0.760}	&\textbf{1.237} & \textbf{2.344}	& \textbf{0.118}  \\  
     & empty map & \textit{transfer}                       & 0.239 &2.385	& 5.152	& 0.481\\
     & empty map & \textit{finetune}             & 0.460	& 1.654	& 3.419	& 0.291\\
     (map: Vector) & MapTR \cite{MapTR} & \textit{transfer} & 0.302	&2.269	&4.863	&0.433 \\
     & MapTR \cite{MapTR} & \textit{finetune} &0.499	&1.670	&3.472	&0.273 \\
     
     \midrule  

     \multirow{3}{*}{ViP3D \cite{gu2023vip3d}} & ground truth & default                   &0.034	&3.540	&5.943	&0.432\\
                                                                           &  empty map & \textit{transfer}                  &0.033	&3.540	&5.943	&0.432\\
     (map: Vector)                                                                      &  empty map & \textit{finetune}        &\textbf{0.040}	&\textbf{3.277}	& \textbf{5.589} &	\textbf{0.404}\\
     \midrule  
     \multirow{3}{*}{UniAD \cite{uniad}} &  online map \cite{uniad} & default & 0.117	&\textbf{1.842}	&3.258	&\textbf{0.228}  \\
                                                                           & empty map & \textit{transfer}               &0.112	&1.908	&3.441	&0.247 \\
      (map: Raster)                                                       & empty map & \textit{finetune}     & \textbf{0.118}	& 1.844	&\textbf{3.250}	&\textbf{0.228}\\
    \bottomrule
    \end{tabular}
    }
  \label{tab:with_wo_map}
\end{table}

%% file: tabs/my_real_world_inputs.tex
\begin{table*}[h]
\centering
\setlength{\tabcolsep}{5.pt} 
    \caption{\textbf{Influence of the perception input type on tracking and forecasting metrics.} Forecasting methods are AgentFormer and LaPred. 
    {\footnotesize$^1$}The past is interpolated when it is incomplete hence {the imperfect MOTA and FP values.\vspace{-.1cm}}}
    \resizebox{\linewidth}{!}{
    \begin{tabular}{@{}c l c c l r r  c c c c c c c c }
    \toprule
     & & \multicolumn{5}{c}{\textit{\textbf{Tracking metrics}}} & \multicolumn{4}{c}{\emph{\textbf{Forecasting metrics for 
 AgentFormer}}} &  \multicolumn{4}{c}{\emph{\eloih{\textbf{Forecasting metrics for LaPred}}}}\\ \cmidrule(lr){3-7}\cmidrule(lr){8-11}\cmidrule(lr){12-15}
     & \multicolumn{1}{l}{\textbf{Perception input}} & \textbf{MOTA} \hspace{-2mm} $\uparrow$ & \textbf{MOTP} \hspace{-2mm} $\downarrow$ & \textbf{FP} \hspace{-2mm} $\downarrow$ & \textbf{FN} \hspace{-2mm} $\downarrow$ & \multicolumn{1}{c}{\textbf{IDS} \hspace{-2mm} $\downarrow$}  & \textbf{mAP$_f$} \hspace{-2mm} $\uparrow$  & \textbf{minADE} \hspace{-2mm} $\downarrow$ & \textbf{minFDE} \hspace{-2mm} $\downarrow$ & \multicolumn{1}{c}{\textbf{MR} \hspace{-2mm} $\downarrow$}  & \eloih{\textbf{mAP$_f$} \hspace{-2mm} $\uparrow$}  & \eloih{\textbf{minADE} \hspace{-2mm} $\downarrow$} & \eloih{\textbf{minFDE} \hspace{-2mm} $\downarrow$} &  \eloih{\textbf{MR} \hspace{-2mm} $\downarrow$} \\ %

     \midrule 
     
     & GT position and identity & \,\,\,0.985\footnotesize$^1$ & 0.000 & \,\,\,155\footnotesize$^1$ & 0 & 0 &  0.343  & 1.885  & 3.979  & 0.359 &  \eloih{0.757}		&	 \eloih{1.213}&	 \eloih{2.316}	& \eloih{0.115}\\
     & GT position + Tracking model \cite{yin2021center} & 0.967 & 0.001 & \,\,\,180 & 0 & 165 & 0.317  &  1.934 & 4.020  &  0.363 &  \eloih{0.730}	&  \eloih{1.353}	&   \eloih{2.513}	& \eloih{0.125} \\

     \midrule
     \multirow{4}{*}{\rotatebox[origin=c]{90}{\parbox[c]{1cm}{Camera-\\based}}} & MUTR3D R50 \cite{zhang2022mutr3d} (CVPRW'22)        &0.170 & 0.607 & 1828 & 6696 & 27 & 0.042  &  3.993 & 7.151  & 0.463 & \eloih{0.152}		&	\eloih{2.237}	& \eloih{3.554}	&\eloih{0.178} \\

    & MUTR3D R101 \cite{zhang2022mutr3d} (CVPRW'22)                                                                          & 0.213 & 0.550 & 1373 & 6727 & 11 &0.055 & 3.480  & 6.286  &  0.449 & \eloih{0.198}		&	\eloih{1.892}	&\eloih{3.043} &	\eloih{\textbf{0.150}}\\
     
     & ViP3D Det\&Track \cite{gu2023vip3d} (CVPR'23)                                                                           & 0.145 & 0.636 & 1855 & 6947 & 3 & 0.056  &  2.416 &  4.404 & \textbf{0.353} & \eloih{0.142}	&	 \eloih{2.044}	& \eloih{3.173}	& \eloih{0.155} \\
     
     & UniAD Det\&Track \cite{uniad} (CVPR'23)                                                                                 & 0.195 & 0.471 & \textbf{1199} & 7076 & 20 & 0.069  & 2.530  & 4.613  & 0.384 & \eloih{0.180}		&	 \eloih{2.142}	& \eloih{3.424}	& \eloih{0.169}\\

       \midrule

    \multirow{3}{*}{\rotatebox[origin=c]{90}{\parbox[c]{0.9cm}{LiDAR-based}}} & MegVii \cite{zhu2019class}+AB3DMOT \cite{Weng2020_AB3DMOT} (IROS'20)    & 0.226 & 0.320 & 1657 & 6232 & 79 & 0.089 & 2.356  & 4.412  & 0.382  & \eloih{0.227}		&	\eloih{2.143}	&\eloih{3.561}	&\eloih{0.168} \\
    
     & CenterPoint \cite{yin2021center} (CVPR'21)                              & \textbf{0.348} & \textbf{0.244} & 1622 & \textbf{5090} & {5} &  \textbf{0.112} & \textbf{2.102}   &  \textbf{4.354}&  0.413 & \eloih{0.285}		&	\eloih{\textbf{1.596}}&	\eloih{\textbf{2.815}}&	\eloih{0.151} \\
     
     & VoxelNext \cite{chen2023voxelnext} (CVPR'23)                            & 0.328 & 0.263& 1639 & 5283 & \textbf{2} & 0.096   & 2.134  & 4.409  &  0.426 & \eloih{\textbf{0.317}}	&		\eloih{1.669}	&\eloih{2.914}	&\eloih{0.166} \\

    \bottomrule
    \end{tabular}
    }
  \label{tab:real_world_results}
  \vspace{-3mm}
\end{table*}

%% file: tabs/det_track_noise.tex
\begin{figure}[t]
\centering
\vspace{-0.25cm}
\begin{subfigure}{0.49\linewidth}
 \caption*{AgentFormer}
\includegraphics[width=\textwidth]{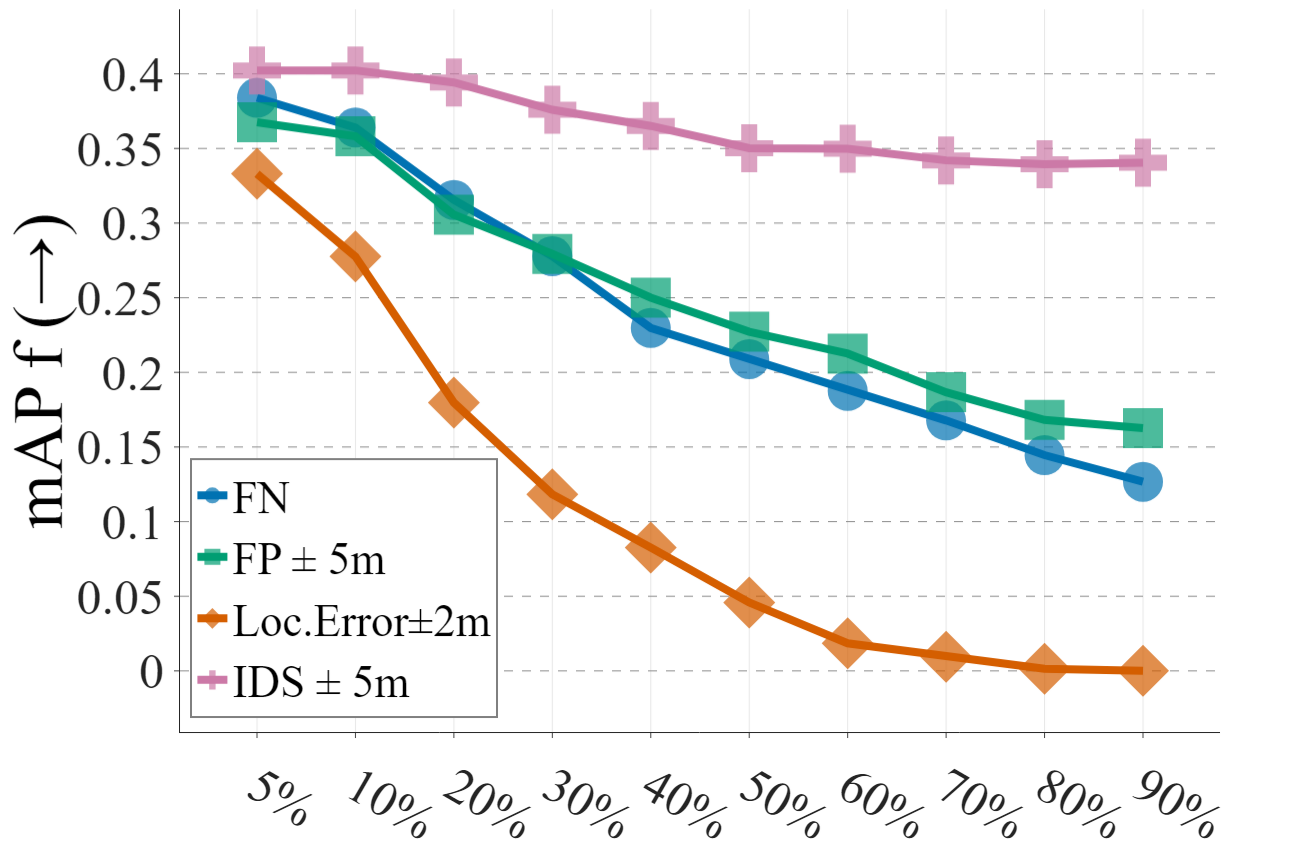} \hspace{-2mm}
\end{subfigure}
\begin{subfigure}{0.49\linewidth}
 \caption*{LaPred}
\includegraphics[width=\textwidth]{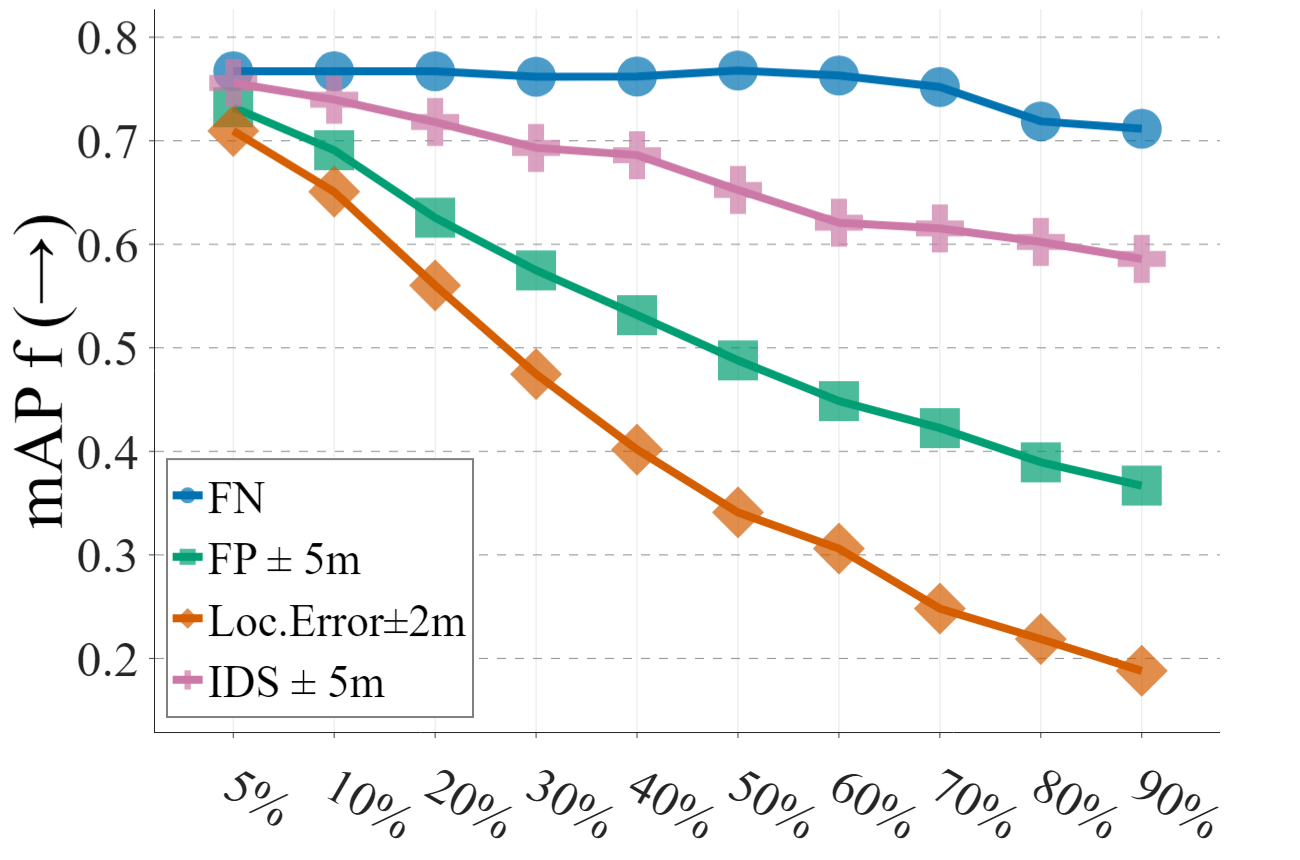} \hspace{-2mm}
\end{subfigure}
\centering
\caption{\textbf{Impact of controlled input errors}
. Forecasting performance (mAP$_f$) {under different proportions of detection and tracking errors} ($x$-axis); We simulate {misdetections (FN, in blue)}, {false detections (FP@5meters, in green)}, {localization errors (Loc. Error@2meters in orange)} and {tracking errors (IDS@5meters, in pink)} in the past trajectories.}
\label{fig:det_track_noise}
\end{figure}

%% file: tabs/training_with_noise.tex
\begin{table}[t]
\centering
    \caption{{\textbf{Impact of finetuning}. LaPred \cite{kim2021lapred} pretrained on GT with data augmentation (DA) is finetuned by either {randomly shifting GT past} trajectories or using results from real-world tracking methods, for 30 epochs with a learning rate 2x slower than the default one (i.e., 5e-5). The finetuning is conducted on the train set and tested on the validation set. Results are relative to training with GT.}\vspace{-0.1cm}}
    
    \resizebox{\linewidth}{!}{
    \begin{tabular}{@{}l l c @{\hspace{2mm}} c @{\hspace{2mm}} c @{\hspace{2mm}} c@{}}
    \toprule
    \textbf{{Finetuning input}} & \eloih{\textbf{Testing input}} & \eloih{\textbf{mAP${_f}$ } \hspace{-2mm} $\uparrow$} & \eloih{\textbf{minADE} $\hspace{-1mm} \downarrow$}  & \eloih{\textbf{minFDE} \hspace{-2mm} $\downarrow$}  & \eloih{\textbf{MR} \hspace{-2mm} $\downarrow$} \\

     \midrule  
     \multirow{4}{*}{{No finetuning}} & \eloih{GT}  & \eloih{0.760} & \eloih{1.237} & \eloih{2.344} & \eloih{0.118} \\  
       &  \eloih{VoxelNext}                                     &  \eloih{0.317} & \eloih{1.669} & \eloih{2.914} & \eloih{0.166}\\ 
       &  \eloih{UniAD}                                        &\eloih{0.180} &\eloih{2.142} &\eloih{3.424} & \eloih{0.169} \\  
       & \eloih{MUTR3D R101}                                      &\eloih{0.198} & \eloih{1.892} & \eloih{3.043} & \eloih{0.150}\\ 

    \midrule   
     \multirow{3}{*}{{{70\% GT\,+\,30\% DA}} } & \eloih{VoxelNext}         &-0.007	&+0.186	&+0.056	&+0.005 \\ 
      & \eloih{UniAD}                                           &+0.007  &-0.166 &-0.324 &-0.015 \\ 
      &  \eloih{MUTR3D R101}                                    &+0.009 & -0.034 &-0.191 &-0.004\\ 
     \midrule   
     \eloih{VoxelNext}  & \eloih{VoxelNext}                            &\eloih{+0.022} & \eloih{-0.142} & \eloih{-0.283} & \eloih{-0.023} \\ 
     \eloih{UniAD} & \eloih{UniAD}                                     &\eloih{+0.018} & \eloih{-0.534} & \eloih{-0.782} & \eloih{-0.020} \\ 
     \eloih{MUTR3D R101} &  \eloih{MUTR3D R101}                        &\eloih{+0.010} & \eloih{-0.135} & \eloih{-0.294} & \eloih{-0.002} \\
    \bottomrule
    \end{tabular}
    }
  \label{tab:train_imperfections}
\end{table}

%% file: tabs/ego_car.tex
\definecolor{myMUTR3D-R50}{RGB}{0, 114, 178}
\definecolor{myMUTR3D-R101}{RGB}{0, 158, 114}
\definecolor{myViP3D}{RGB}{212, 94, 0}
\definecolor{myMegVii-AB3DMOT}{RGB}{239, 201, 66}
\definecolor{myUniAD}{RGB}{204, 121, 167}
\definecolor{myCenterPoint}{RGB}{86, 179, 232}
\definecolor{myVoxelNext}{RGB}{0, 41, 172}
\definecolor{myGT-Tracking}{RGB}{255, 128, 14}
\definecolor{myGT}{RGB}{171, 171, 171}

\begin{figure}[t]
\centering
\begin{subfigure}{0.48\linewidth}
\includegraphics[trim={0.0cm 2.0cm 2.0cm 2.0cm},clip,width=\linewidth]{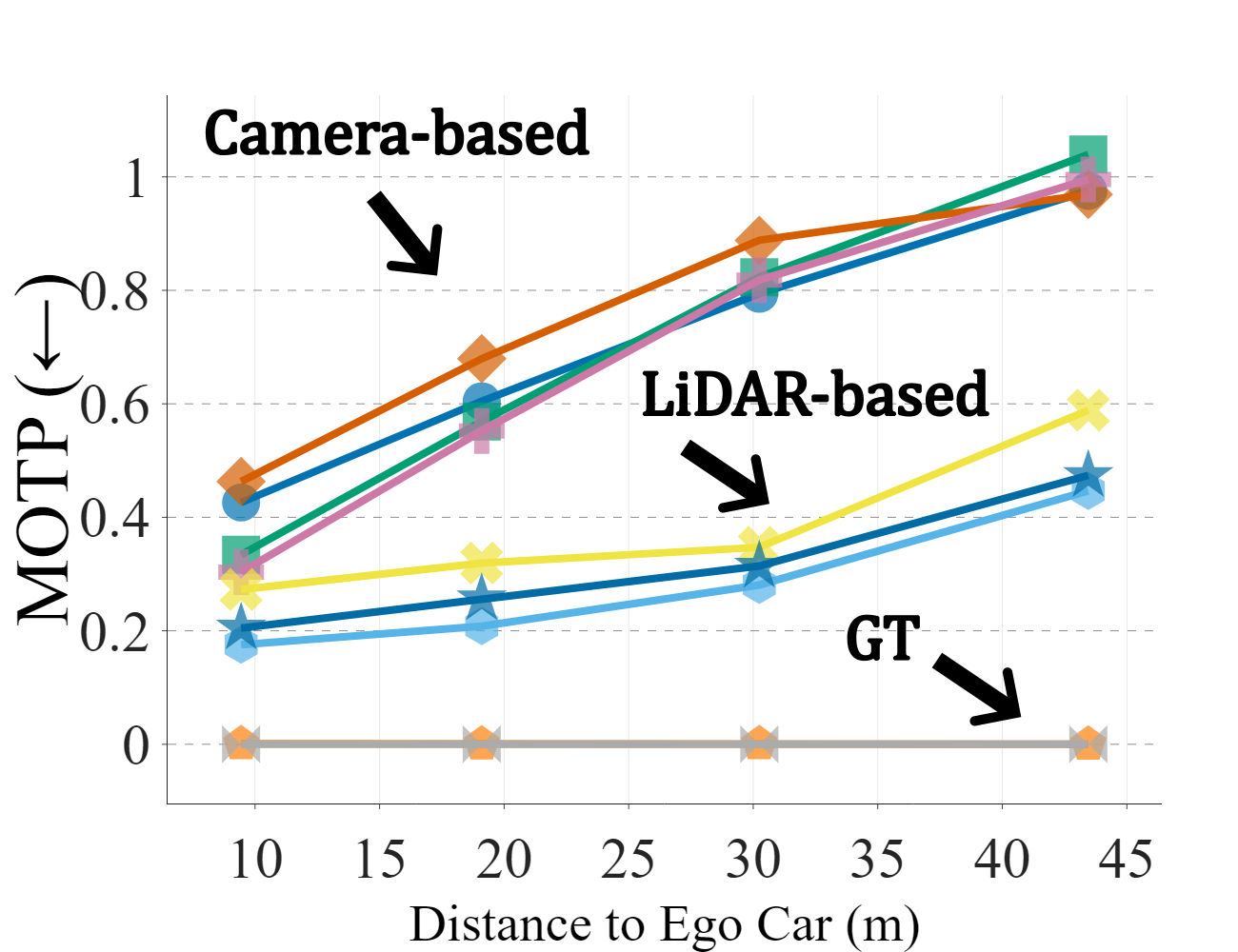}
\end{subfigure}
\hfill
\begin{subfigure}{0.48\linewidth}
\includegraphics[trim={0.0cm 1.6cm 1.8cm 1.8cm},clip,width=\linewidth]{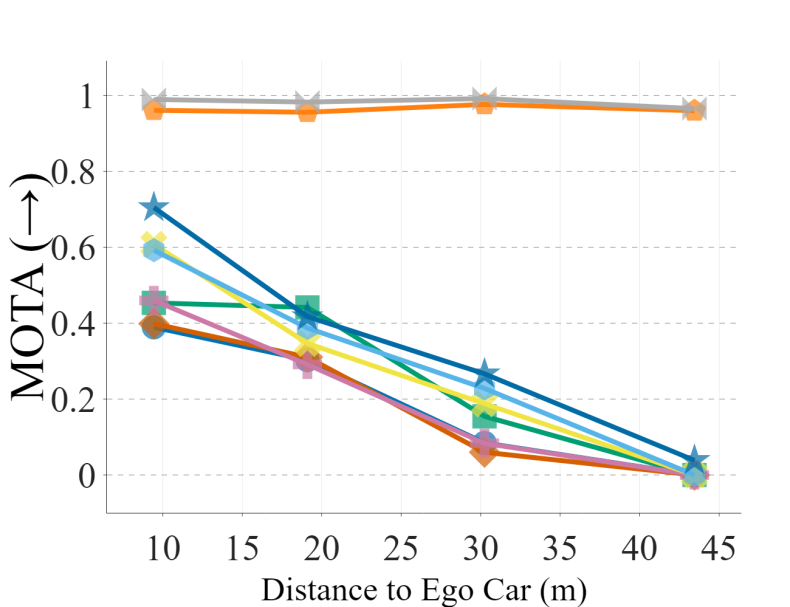}
\end{subfigure}
\begin{subfigure}{0.39\linewidth}
\includegraphics[trim={0.0cm 2.0cm 2.0cm 2.0cm},clip,width=\linewidth]{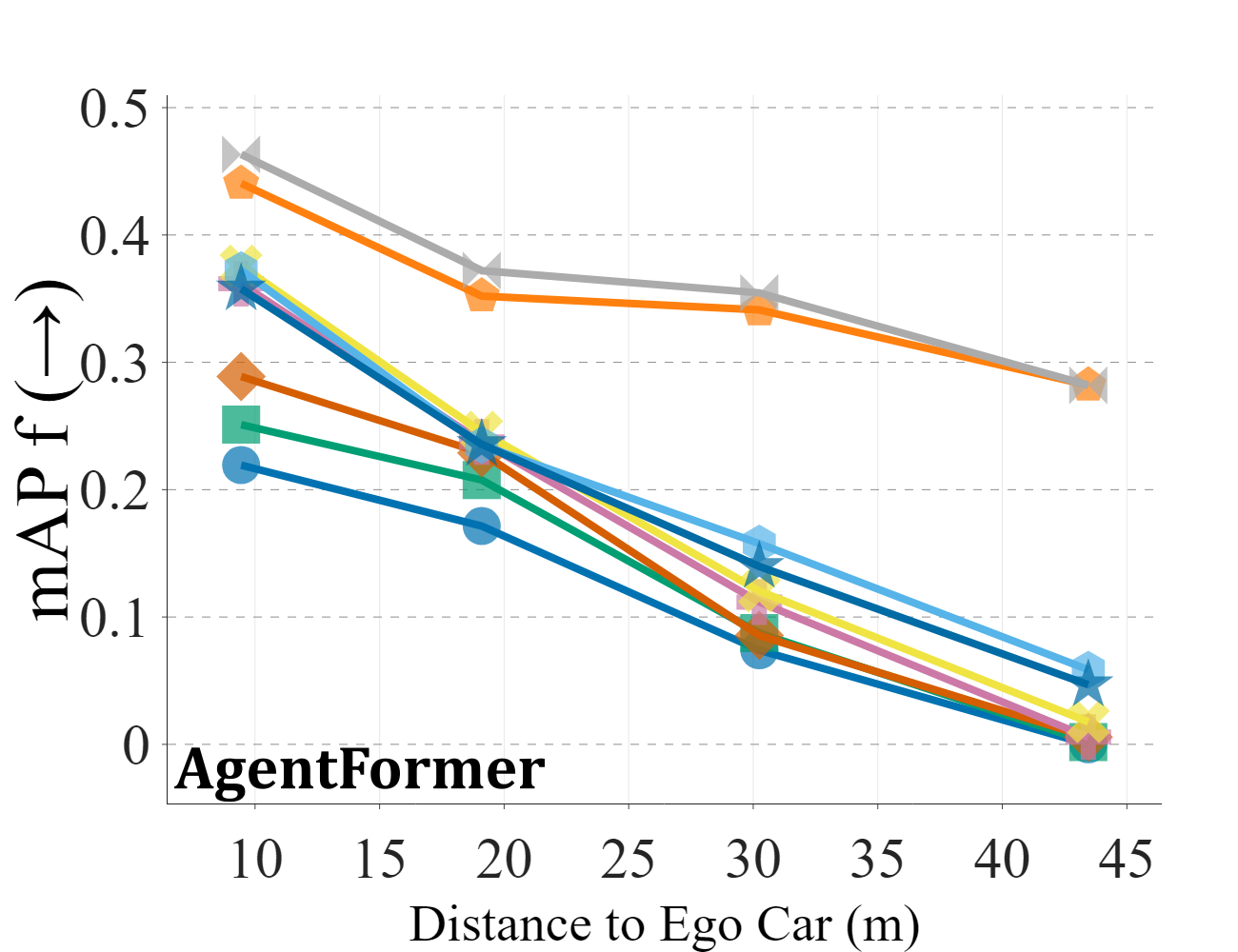}
\end{subfigure}
\hfill
\begin{subfigure}{0.39\linewidth}
\includegraphics[trim={0.0cm 2.0cm 2.0cm 2.0cm},clip,width=\linewidth]{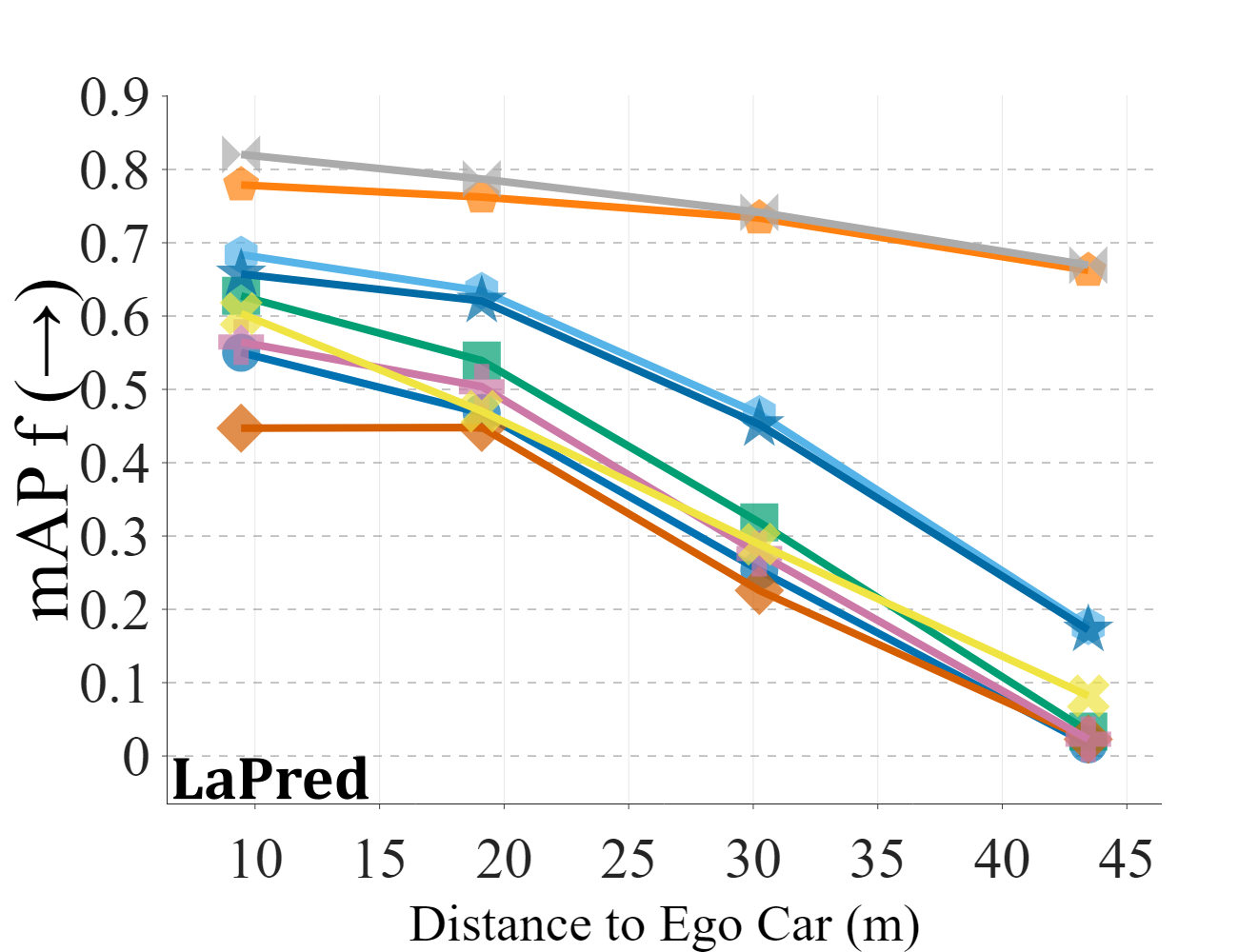}
\end{subfigure}
\hfill
\begin{subfigure}{0.19\linewidth}
\includegraphics[trim={0.0cm 0.0cm 0.0cm 0.0cm},clip,width=\linewidth]{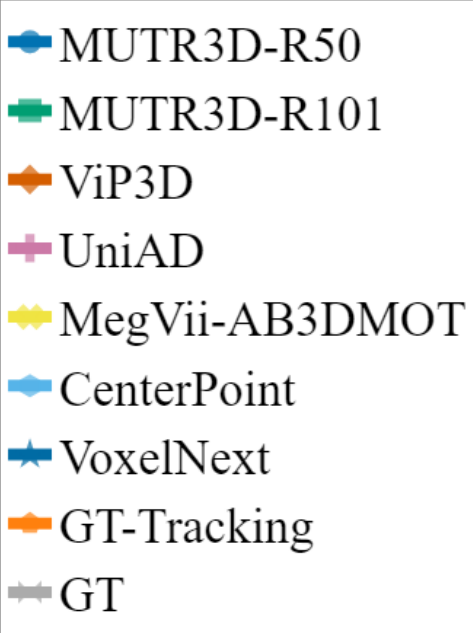}
\end{subfigure}
\caption{\textbf{Impact of agent-ego distance.} Tracking and forecasting performances w.r.t. agent-ego distance ($x$-axis in meters) for tracking methods: (camera-based) {MUTR3D-R50}, {MUTR3D-R101}, {ViP3D}, {UniAD}; (LiDAR-based) {MegVii-AB3DMOT}, {CenterPoint}, {VoxelNext}; {GT-Tracking} and {GT}.
}
\label{fig:dist_ego_car}
\end{figure}